\documentclass[conference]{IEEEtran}
\IEEEoverridecommandlockouts
\usepackage{cite}
\usepackage{amsmath,amssymb,amsfonts}
\usepackage{algorithmic}
\usepackage{graphicx}
\usepackage{algorithmic}
\usepackage{algorithm}
\usepackage{multirow}
\usepackage{xcolor}
\usepackage{colortbl}
\usepackage{setspace}
\usepackage{wrapfig}
\usepackage{color}
\usepackage{amsmath,amsfonts,amssymb}
\usepackage{textcomp}
\usepackage{longtable}% for long tables
\usepackage{booktabs}
\usepackage{ulem}
\usepackage{amsmath, nccmath}
\usepackage{lipsum}
\def\Reel{\textrm{I\kern-0.21emR}} %idem mais sans doublage de la boucle
\usepackage{xcolor}
\def\BibTeX{{\rm B\kern-.05em{\sc i\kern-.025em b}\kern-.08em
    T\kern-.1667em\lower.7ex\hbox{E}\kern-.125emX}}
\begin{document}

\title{From Weakly Supervised Learning to Biquality Learning: an Introduction 
%and perspectives
%based on three axis  : Adaptability, Quality, Quantity
}

%%%%%%%%%%%%%%%%%%%
\author{
\IEEEauthorblockN{Pierre Nodet}
\IEEEauthorblockA{
\textit{Orange Labs} \\
\textit{AgroParisTech, INRAe}\\
46 av. de la République\\
Châtillon, France}
\and
\IEEEauthorblockN{Vincent Lemaire}
\IEEEauthorblockA{\textit{Orange Labs} \\
2 av. P. Marzin\\
Lannion, France}
\and
\IEEEauthorblockN{Alexis Bondu}
\IEEEauthorblockA{\textit{Orange Labs} \\
46 av. de la République\\
Châtillon, France}
\and
\IEEEauthorblockN{Antoine Cornuéjols}
\IEEEauthorblockA{\textit{UMR MIA-Paris}\\ 
\textit{AgroParisTech, INRAe}\\
\textit{Universit\'e Paris-Saclay}\\
16 r. Claude Bernard\\
Paris, France}
\and
\IEEEauthorblockN{Adam Ouorou}
\IEEEauthorblockA{\textit{Orange Labs} \\
46 av. de la République\\
Châtillon, France}
}

\maketitle

\begin{abstract}
The field of Weakly Supervised Learning (WSL) has recently seen a surge of popularity, with numerous papers addressing different types of
``supervision deficiencies''. In WSL use cases, a  variety  of  situations exists where the  collected ``information'' is imperfect.
 The paradigm of WSL attempts to list and cover these problems with associated solutions. 
In this paper, we review the research progress on WSL with the aim to make it as a brief introduction to this field. We present the three axis of WSL cube and an  overview  of  most  of  all  the  elements  of  their  facets. We propose three measurable quantities that acts as coordinates in the previously defined cube namely: Quality, Adaptability and Quantity of information. Thus we suggest that Biquality Learning framework can be defined as a plan of the WSL cube and propose to re-discover previously unrelated patches in WSL literature as a unified Biquality Learning literature.
\end{abstract}


%%%%%%%%%%%%%%%%%%%%%%%%%%%%%%%%%%%%%%%%%%%%%%%%%%%%%%%%%%%%%%%%%%%%%%%%%%%%%%%%%%%%%%%%%%%%%%%%%%%%%%%%%%%%%%%%%%%%%%%%%%%%%%%%%%%%%%%%%%%%%%%%%%%%%%%%%%%%%%%%%%%%%%%%%%%%%%%%%%%%%%%%%%%%%%%%%%%%%%%%%%%%%%%%%%%%%%%%%%%%%%%

\section{Introduction}
\label{intro}

%Machine learning is mostly about fitting a problem into two different approaches: \textit{unsupervised} or \textit{supervised} learning. The first allows discovering new tendencies, rules or categories among input data with limited prior knowledge. The second allows expanding prior knowledge on data, with the idea that a good way of predicting future examples is observing past ones. These examples are typically inputted with an associated label defined per individual. Supervised learning techniques have shown impressive results on various applications, especially in the case of large datasets.

%Classification could be done either with or without labels using respectively supervised or unsupervised algorithm. 
%It is obvious that the resulting performances, in a supervised meaning, will be more or less interesting depending of  a 'learning cost' as presented in Figure \ref{classofclass}. In this figure Sugiyama \cite{SugiyamaTalkIdiap} indicates that an interesting goal could be to obtain a high accuracy while spending a low learning, or labeling, cost.

\begin{figure}[h]
\centering
\includegraphics[width=\linewidth]{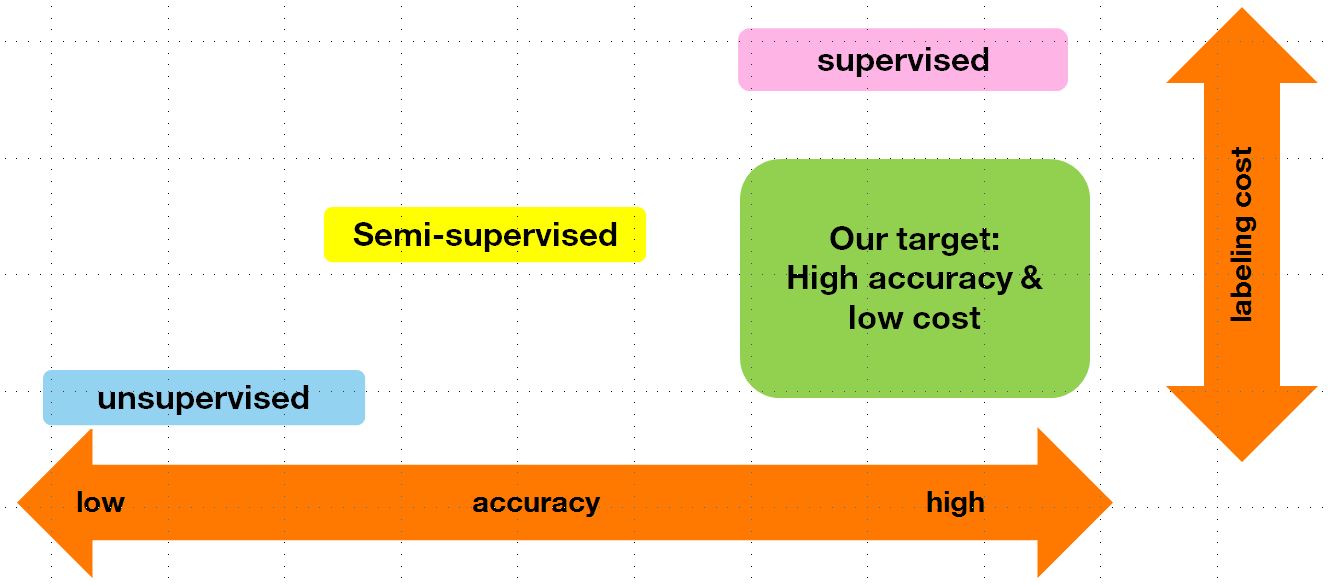}
\caption{Classification of Classification from \cite{SugiyamaTalkIdiap}}
\label{classofclass}
\end{figure}

In the field of machine learning, the task of classification can be performed by different %types of AO: pas necessaire a mon avis 
approaches depending on the level of supervision of training data. 
As shown in Figure \ref{classofclass}, unsupervised, weakly supervised and supervised approaches form a continuum of possible situations, starting from %with 
the absence of ground truth and ending with complete and perfect ground truth. For the most part, the accuracy of the models learned increases as the level of supervision of data increases. 
Additionally, the level of supervision of a dataset can be increased in return for a labelling cost.  
In \cite{SugiyamaTalkIdiap}, the authors indicate that an interesting goal could be to obtain a high accuracy while spending a low labeling, cost.

In \textit{Weakly Supervised Learning} (WSL) use cases (e.g. fraud detection) a variety of situations exist where the collected ground truth is imperfect.  
In this context, the collected labels may suffer from bad quality, non-adaptability (defined in Section \ref{3concepts}) or even insufficient quantity.
%In some applications as fraud detection, the question is also not only to have labels but the quality, adaptability or quantity of labels. Getting exact ground truth labels can be difficult in this context, as it could be in many others.  There are multiple ways of getting target values. 
For instance, automatic labeling system could be used without any real guarantee that the data is complete, exhaustive and trustworthy.
%Open datasets or automatic collection could be used, with no real guarantee of the data being complete, exhaustive and trust-worthy. 
%Manual labeling is also an alternative, but getting appropriate knowledge from an expert implies a cost, where sufficient amount of information are often unaffordable. 
Alternatively, manual labelling is also problematic in practice as obtaining labels from an expert is costly and the availability of experts is often limited.
Consequently, there are many real-life situations where imperfect ground truth must be used because of some practical considerations such as
%for a variety of practical reasons - e.g. 
cost optimization, expert availability, difficulty to certainly choose each label.

This general problem of {supervision deficiency} has attracted a recent focus in the literature. The paradigm of \textit{Weakly Supervised Learning} attempts to list and cover these problems with associated solutions. The work of Zhou in \cite{zhou2017} is a first successful effort {to synthesise} this domain. In this paper, the objective is threefold: (i) to suggest another view of WSL, (ii) {to propose a larger and updated taxonomy compared to} \cite{zhou2017}, and then (iii) to highlight a new emergent view of a part of the WSL, namely the biquality learning.

%\textcolor{magenta}{Do we need this paragraph ? : The rest of this paper is organized as follow. In Section II we present the three axis of the Weakly supervised Learning cube and an overview of most of all the elements of their facets. The Section III gives additional elements which have to be taken in consideration at the crossroad of these three axes or when dealing with Weakly learning problems. Then, knowing all the elements given in sections II and III, the Section \ref{3concepts} suggest 3 key concepts which help at summarizing WSL:  Quantity, Quality and Adaptability.
%The use,  in Section V, of these 3 concepts allow to raise  links between some learning frameworks jointly used in WSL as in Biquality Learning. The section VI then give existing works examples of Biquality Learning. Finaly the Section VII concludes this paper.}
The rest of this paper is organized as follows. In Section II, we present the three axis of the Weakly supervised Learning cube and an overview of most of all the elements of their facets. Section III gives additional elements which have to be taken in consideration at the crossroad of these three axes or when dealing with Weakly learning problems. 
%Then, knowing all the elements given in sections II and III, the 
Section \ref{3concepts} suggests 3 key concepts which help at summarizing WSL:  Quantity, Quality and Adaptability. In Section V, these 3 concepts are used to raise links between some learning frameworks jointly used in WSL as in Biquality Learning. Section VI then give existing works examples of Biquality Learning. Finally the Section VII concludes this paper.

%%%%%%%%%%%%%%%%%%%%%%%%%%%%%%%%%%%%%%%%%%%%%%%%%%%%%%%%%%%%%%%%%%%%%%%%%%%%%%%%%%%%%%%%%%%%%%%%%%%%%%%%%%%%%%%%%%%%%%%%%%%%%%%%%%%%%%%%%%%%%%%%%%%%%%%%%%%%%%%%%%%%%%%%%%%%%%%%%%%%%%%%%%%%%%%%%%%%%%%%%%%%%%%%%%%%%%%%%%%%%%%

\begin{figure*}[!h]
\centering
\includegraphics[width=0.72\linewidth]{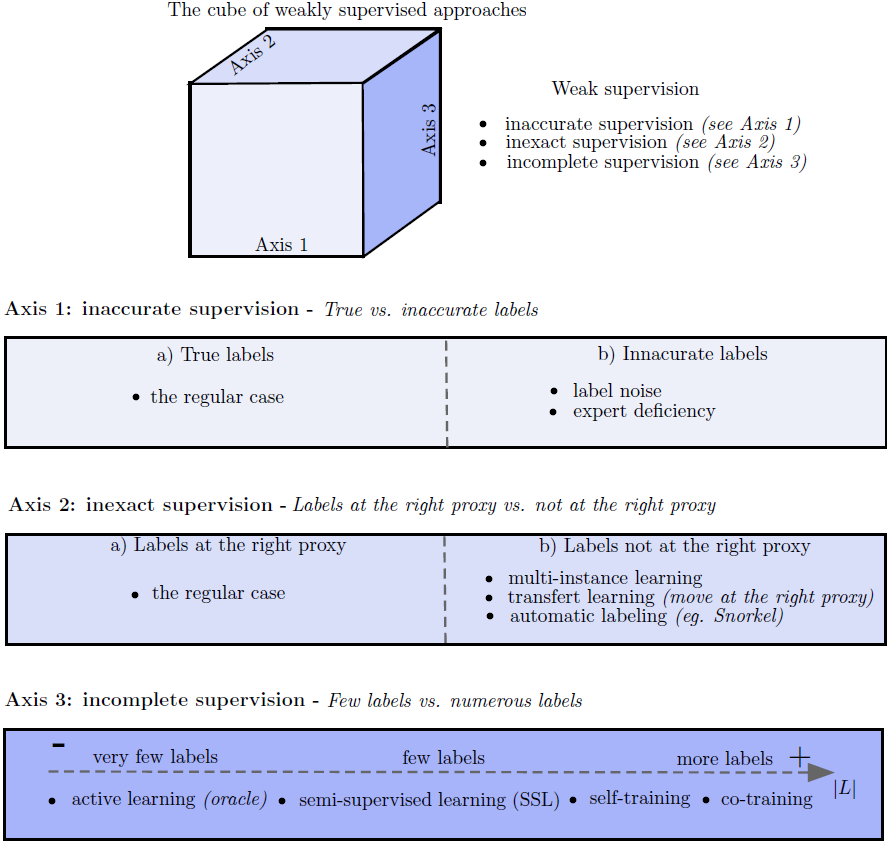}
\caption{Taxonomy: an  attempt - The big picture}
\label{bigpicture}
\end{figure*}

%%%%%%%%%%%%%%%%%%%%%%%%%%%%%%%%%%%%%%%%%%%%%%%%%%%%%%%%%%%%%%%%%%%%%%%%%%%%%%%%%%%%%%%%%%%%%%%%%%%%%%%%%%%%%%%%%%%%%%%%%%%%%%%%%%%%%%%%%%%%%%%%%%%%%%%%%%%%%%%%%%%%%%%%%%%%%%%%%%%%%%%%%%%%%%%%%%%%%%%%%%%%%%%%%%%%%%%%%%%%%%%
\section{The different ways of looking at weak supervision}

%The Taxonomy, an  attempt in  this paper, suggested is organized as a 'tree' presented in Figure \ref{bigpicture}. 
%The actual section presents progressively the different nodes and leaves of this tree.

The taxonomy proposed in this paper is organised in the form of a ``cube" and is presented in Figure \ref{bigpicture}.
This section progressively presents the differences between weakly supervised approaches by going through the axes of this cube.

%%%%%%%%%%%%%%%
%\subsection{Strong \textit{vs.} weak supervision}
%\label{strongvsweak}

%First of all the main distinction between ``strong versus weak'' learning lies in two aspects : supervision and labels. The weakness concerning the supervision is presented here, the one concerning the labels is presented in the next subsection (Section \ref{labels}).
%Three main types of weak ‘supervision’ exist:

First of all, a distinction must be made between \textit{strong} and \textit{weak} supervision. 
On the one hand, \textit{strong} supervision correspond to the regular case in \textit{machine learning} where the training examples are expected to be exhaustively labelled with true labels, i.e. without any kind of corruption or deficiency.   
On the other hand, \textit{weak} supervision means that the available ground truth is imperfect, or even corrupted. The WSL field aims to address a variety of supervision deficiencies which can be categorized in a ``cube'' along the following three axes as illustrated on Figure~\ref{bigpicture}: inaccurate labels (Axis 1), inexact labels (Axis 2), incomplete labels (Axis 3).

These three axes are detailed in the rest of this section and constitute the proposed taxonomy. The reader may note that the boundaries between these axes are not hard: i.e a part could be moved from an axis to another or belong to two axes, this is a  suggestion.

%%% Résumé de zoo - 

\subsection{{\bf Axis 1}: Inaccurate Supervision - True Labels \textit{vs.} Inaccurate Labels}
\label{axis1}

%00 Inaccurate supervision: Another 

%A common case when dealing with real tasks is the lack of trust in the data sources. The values used as a target for learning, also called labels or classes, could be wrong, due to a multiplicity of factors. 
%Results of supervision rely on the quality of labels, and having a \textit{weak} knowledge about the final target naturally affects performances\footnote{These errors would be referred as \textit{noise} or \textit{corruption}. This topic is the closest to the weaknesses which will described into the biquality learning (see Section XXX)}.

Lack of confidence in data sources is a frequent issue when it comes to real-life use cases. The values used as learning targets, also called labels or classes, can be incorrect due to many factors. 
%\textcolor{gray}{Strong supervision relies on high-quality labels. On the contrary, weak supervision is due to a partial knowledge of the target to be learned, which naturally affects the classifier's performance\footnote{These errors are called label noise or label corruption. Such supervision deficiencies are addressed by bi-quality learning (see section XXX).}.\textcolor{red}{Redondant avec l'intro de la Section 2}
%0000 T
%The weakness or deficiencies on the labels could be due to two main issues: (i) Inaccurate or imprecise labels, (ii)
%True labels but incomplete supervision (incomplete information).
%\paragraph{Inaccurate or imprecise labels:}

In practice, a variety of situations can lead to inaccurate labels: 
(i) a label can be assigned to a ``bag of examples " such as a bunch of keys. In this case, at least one of the examples in the keychain actually belongs to the class indicated by the label. Multi-instance learning \cite{yang2005review,zhou2006multi,foulds2010review,Carbonneau_2018} is an appropriate technique to deal with this type of of learning task. 
(ii) a label may not be ``guaranteed" and may be noisy. In theory, the learning set should be labeled in a way that is unbiased with respect to the concept to be learned.  However, the data used in real-world applications provide an imperfect ground truth that does not %perfectly %AO: imperfect l'exprime deja 
match the concept to be learned.  
%a label could be not ‘guaranteed’the labels are noisy. In theory, the dataset used for training is supposed to represent a subset of the ground truth. However, data in real-world applications rarely correspond perfectly to reality. 
As defined in \cite{hickey_NoiseModellingEvaluating_1996}, noise is “anything that obscures the relationship between the features of an instance and its class”. According to this definition, every \textit{error} or \textit{imprecision} in a attribute or label is considered as noise, including human deficiency.
Noise is not a trivial issue because the origin is never clearly obvious. 
In practical cases, this leads to troubles into evaluating existence and strength level of noise into a dataset. 
Frenay et al. in \cite{frenay_ClassificationPresenceLabel_2014} provide a good overview of noise sources, impact of labeling noise, types of noise and dependency to noise.
Below is a non-exhaustive list of common ways to learn a model in the presence of labeling noise\footnote{Note: the number of articles published on this topic has exploded in recent years.}:

\begin{itemize}
        \item %if the noise level is marginal, use a standard learning algorithm that is natively robust to label noise 
        in case of mariginal noise level, a standard learning algorithm that is natively robust to label noise, is used 
        \cite{lebaher2020,nettleton_StudyEffectDifferent_2010,folleco_IdentifyingLearnersRobust_2008,zhu_ClassNoiseVs_2004}; 
        \item use a loss function which solves theoretically (or empirically) the problem in case of (i)  noise completely at random\footnote{defined in Section \ref{quality}.} \cite{charoenphakdee_symmetric_2019}; or (ii) class dependent noise \cite{Hendrycks2018,Xia2019AreAP}. In most cases, this type of approach is known in the literature as ``Robust  Learning to  Label  noise  (RLL)";
        \item model noise to assess the quality of each label (requires assumptions on noise) \cite{Sukhbaatar2014TrainingCN};
        \item {enforce consistency between the model's predictions and the proxy labels \cite{Reed2015TrainingDN}};
        \item clean the training set by filtering noisy examples \cite{sun_identifying_2007,malossini_detecting_2006, miranda_use_2009,matic_computer_1992,van_hulse_knowledge_2009};
        \item trust a subset of data provided by the user, in order to learn a model at once on trusted examples (without label noise) and untrusted ones \cite{Hendrycks2018,Hendrycks2019,nodet2021importance}. 
\end{itemize}

Another kind of ''noise'' appears when  each training example is not  equipped with a single true label but with a set of candidate labels that contains the true label. 
To deal with this kind of training examples, Partial Label Learning (PLL) has  been proposed \cite{PLL1} (also called ambiguously labeled learning). It has attracted attention as for example in the algorithms IPAL \cite{zhang2015solving}, PL-KNN \cite{PLL1}, CLPL \cite{cour2011learning} and PL-SVM \cite{nguyen2008classification} or when suggesting semi-supervised partial label learning as in  \cite{ijcai2019-521}. This setting is motivated, for example, by  common scenario in many image and video collections, where only partial access to labels is available.  The goal is to learn a classifier that can disambiguate the partially-labeled training instances, and generalize to unseen data \cite{JMLR:v12:cour11a}.

 %We place here PLL since it could be view as noisy labels 
    
%\textcolor{magenta}{Inaccurate supervision, could also be due when few labels are available in this case the litterature present. This point named 'incomplete supervision' is detailed in Section \ref{axis3}}.

\subsection{{\bf Axis 2}: Inexact Supervision - Labels at the Right Proxy \textit{vs.} not at the Right Proxy}

%OO Inexact supervision: 
%\textcolor{gray}{
%The second type of supervision deficiency, described in \cite{zhou2017}, is more specific that the previous one.}\textcolor{red}{why specific?} 

The second axis describes \textit{inexact labeling} which is orthogonal to the first type of supervision deficiency - i.e. inexact labeling and noisy labeling may coexist.
Here, the labels are provided not at the right proxy, which corresponds to one (or possibly a mixture) of the following situations:

%Here, the labels can be trustful and never missing, but there are provided not at the proper format, not as exact as
%desired or not at the right proxy\footnote{Note (to check if this note is useful or not): we considered the term ``proxy'' in two sens int this paper.

\begin{itemize}
    \item \textit{Proxy domain}: the target domain differs between the training set and the test set. For instance, it could be learning to discriminate ``panthers" from other savanna animals based one ``cats" and ``dogs" labels. Two cases can be distinguished: (i) training labels are available in another target domain than test labels (ii) or training labels are available in a sub-domain that belongs to the original target domain. Domain transfer \cite{duan2012domain} or domain adaptation \cite{ben2010theory} are clearly suitable techniques to address these learning tasks.
    %proxy is the native domain of a initial label. Thus it could not be adapted (learning to classify cats using dogs labels).
    
    \item \textit{Proxy labels}: some unlabeled examples are automatically labeled, either by a rule-based system or by a learned model, in order to increase the size of the training set. This kind of labels are called proxy labels and can be considered as coming from a proxy concept close to the one to be learned. Only the true labels stand for the ground truth. The way proxy labels are used varies depending on their origin. In the case where proxy labels are provided by the classifier itself without any additional supervision, the self-training (ST) \cite{ennaji2012self}, the co-training (CT) and their variants attempt to improve the learned model by including proxy-labels in the training set as regular labels. Other approaches exploits the confidence level of the classifier to produce soft-proxy-labels, and then exploit it as weighted training examples \cite{torgo20182nd}. In the case where proxy labels are generated by a rule-based system, the quality of labels depends on the experts knowledge which is manually inputted into the rules. Ultimately, a classifier learned from such labels can be considered as a means of smoothing the set of rules, allowing the end-user to score any new example. Some recent automatic labeling system offer an intermediate way that mixes rule-based systems and machine learning approaches (MIX) \cite{ratner2020snorkel,varma2018snuba}. 
    
    \item \textit{Proxy individuals}: the statistical individuals are not equally defined between the training set and the test set. For instance, it could be learning to classify images based one labels that only describe a parts of the images. Multi-instance learning (MIL) is an other example which consists in learning from labeled groups of individuals. In the literature, many algorithms have been adapted to work within this paradigm \cite{yang2005review,zhou2006multi,foulds2010review,Carbonneau_2018}.

\end{itemize}

\subsection{{\bf Axis 3}: Incomplete Supervision - Few labels \textit{vs.} Numerous}
\label{axis3}

The third axis describes incomplete supervision which consists of processing a \textit{partially labeled} training set.
In this situation, labeled and unlabeled examples coexist within the training set, and it is assumed that there are not enough labeled examples to train a performing classifier. 
The objective is to use the entire training set, including the unlabeled examples, to achieve better classification performance than learning a classifier only from labeled examples.

%One common case of Incomplete supervision is having both labeled (where their 'quantity' is which is insufficient to train a good learner) and unlabeled individuals into the dataset.

%Unlabeled would only mean \textit{missing} in this context. 
%Transforming the problem into a supervised approach would be fairly simple, as we would be able to select only the examples that allows the supervision. 
%Since we would want to use the full amount to maximize performances, this naive approach is not advised if it let a tiny usable subset for learning. In incomplete supervision.

In the literature, many techniques exist capable of processing partially labeled training data, i.e. active learning (AL), semi-supervised learning (SSL), positive unlabeled learning (PUL) or self-training (ST) and Co-Training (CT). At the bottom of the Figure \ref{bigpicture}, we suggest to sort these methods according to the quantity of labeled examples they require. All these approaches are  detailed below.

\subsubsection{Active Learning (AL) \cite{settles2009active}}
Modern supervised learning approaches are known to require large amounts of training examples in order to achieve their best performance. 
%Since these examples are mainly obtained through human experts who manually label examples, the labelling process may have a high cost. 
These examples are mainly obtained by human experts who label them manually, making the labelling process costly in practice. %the labeling process can be costly in practice. 
Active learning~(AL) is a field that includes all the selection strategies that allow one to iteratively build the training set of a model in interaction with a human expert (also called oracle).
The aim is to select the most informative examples to minimize the labelling cost.

Active learning is an iterative process that continues until a labelling budget is exhausted or a predefined performance threshold is reached. 
Each iteration begins with the selection of the most informative example. 
This selection is generally based on information collected during previous iterations (predictions of a classifier, density measures, etc.). 
The selected example is then submitted to the oracle which returns the associated class, and the example is added to the training set ($L$). 
The new learning set is then used to improve the model and the new predictions are used to perform the next iteration.

%In traditional heuristic-based AL the utility measures defined by the active learning strategies in the literature~\cite{settles2009active} differ in their positioning according to a dilemma between the exploitation of the current classifier and the exploration of the training data. 

In conventional heuristic %AL, 
the utility measures used by active learning strategies \cite{settles2009active} differ in their positioning with respect to the trade off %dilemma 
between exploiting the current classifier and exploring training data.
Selecting an unlabelled example in an unknown region of the observation space helps to explore the data, so as to limit the risk of learning a hypothesis that is too specific to the set $L$ of currently labeled examples. %$|L|$.
Conversely, selecting an example in an already sampled region allows to locally refine the predictive model.
%This document is not intended to provide an exhaustive overview of existing AL strategies, the reader may find (i) a more derailed overview in \cite{Aggarwal2014,settles2009active} (ii) a recent benchmark comparisons in \cite{Santos2014,Yang2018,PereiraSantos2019} and (iii) new way to envision the uncertainty in \cite{hllermeier2019aleatoric}.
We do not intend to provide an exhaustive overview of existing AL strategies and refer to~ \cite{Aggarwal2014,settles2009active} for a detailed overview, \cite{Santos2014,Yang2018,PereiraSantos2019} for some recent benchmark and a new way to treat uncertainty in~ \cite{hllermeier2019aleatoric}

%In meta-learning approaches to active learning algorithms within this paradigm are first designed to combine traditional AL strategies with bandit algorithms
Another \textit{meta active learning} paradigm exists, which combines
conventional strategies using bandit algorithms 
\cite{Baram2004,Ebert2012,Hsu2015,Chu2016,Collet2018,Pang2018}. 
%These meta-learning algorithms learn how to select the best AL criterion for any given dataset and adapt it over time as the learner improves. 
These meta-learning algorithms intend to select online the best AL strategy %, i.e. 
according to the observed improvements of the classifier. These algorithms are capable of adapting their choice over time as the classifier improves.
%
%However, all the learning must be achieved within a few examples to be helpful, and these algorithms suffer from a cold start issue. 
However, learning must be done using few examples to be useful and these kind of algorithms suffer from the cold start problem.
In addition these approaches are limited to combine existing AL heuristic strategies.

%Within the meta-learning paradigm, some other algorithms have been developed to learn from scratch an AL strategy on multiple source datasets and transfer it to new target datasets
%
Other meta-active-learning algorithms have been developed to learn an AL strategy starting from scratch, using multiple source of datasets. These algorithms are used by transferring the learned AL strategy to new target datasets \cite{Konyushkova2017,Konyushkova2018,Pang2018b}. 
Most of them are based on modern reinforcement learning methods. 
The major challenge is to learn an AL strategy general enough to automatically control the exploitation/exploration trade-off when used on new unlabeled datasets (which is impossible using heuristic strategies).
A recent evaluation of learning active learning can be found in \cite{Desreumaux2020LearningAL}.

\subsubsection{Semi Supervised Learning (SSL)}

%Semi-supervised learning is not recent and an overview can be seen in
Early work on semi-supervised learning %(SSL) AO: de trop
dates back to the 2000s, an overview of these pioneering papers can be found in  \cite{seeger2000learning,chapelle_semi-supervised_2006,chapelle2009semi,zhu2005semi,zhou2010semi}. %Semi-supervised learning has seen some success.
In the literature, the SSL approaches can be categorized into two groups: 

\begin{itemize}
    %\item algorithms that used unlabeled data as it is as an additional information to the labeled data.
    %{There are four major categories: generative methods, graph-based methods, low density separation methods and disagreement-based methods. - to check, lire zhou, renvoyer vers zhou?}
    
    \item Algorithms that use unlabeled examples unchanged.  In this case, the unlabeled examples are treated as unsupervised information added to the labeled examples. Four main categories exist: generative methods, graph-based methods, low-density separation methods, and disagreement-based methods \cite{zhou2017}.
    
    \item %semi-supervised learning algorithms that produce proxy labels on unlabelled data, which are used as targets together with the labelled data. 
    Semi-supervised learning algorithms that produce proxy labels on unlabeled examples, which are used as targets in addition to the labeled examples.
    %These proxy labels are produced by the model itself or variants of it without any additional supervision; they thus do not reflect the ground truth but might still provide some signal for learning. 
    These proxy labels are produced by the model itself or by its variants%variants of it, 
    without any additional supervision. They are not strictly ground truth, but may nevertheless be useful for learning.
    At the end, these inaccurate labels (see Section \ref{axis1}) can be considered as noisy. 
    %In this second group,  one find Self-Taining, Cotraining, ..., some of which are described just below.
    The rest of this section deals with particular cases of SSL and presents the Postive Unlabeled Learning , the Self Training and the Co-training approaches.
\end{itemize}

{
\subsubsection{Postive Unlabeled Learning (PUL)}
%\vspace{1cm}
Learning from Positive and Unlabeled examples (PUL) is a special case of binary classification and SSL \cite{Bekker_2020}. 
%PUL is the setting where a learner only has access to positive examples and unlabeled data. 
In this particular setting, the unlabeled examples may contain both positive and negative examples with hidden labels.
%This setting fits well with the interest in developing learning algorithms that do not require fully supervised data.
%It differs from one class classification \cite{khan2014one} in that it explicitly incorporates unlabeled examples into the learning process.
These approaches differ from a one-class classification \cite{khan2014one} since they explicitly use the unlabeled examples in the learning process.
%It is related to SSL in that it specializes the semi-supervised setting.
In the literature, the PUL approches can be divided into three groups:  
%Most methods which adress PUL can be divided into  three categories: (
i) the two-step techniques, 
(ii) the biased learning and 
(iii) the class prior incorporation techniques.}

The two-step techniques \cite{liu2003building} consists in: (1) identifying reliable negative examples and optionally generating additional positive examples \cite{TKDE.2006.16};
(2) using supervised or semi-supervised learning approaches which process the positively labeled examples, the reliable negatives examples, and the remaining unlabeled examples;
(3) (when applicable) selecting the best classifier generated in Step~2.
Biased learning approaches consider PU data as fully labeled examples with noisy negative labels. 
At last, class prior incorporation approaches modify standard learning algorithms by applying the mathematics from the SCAR setup (see Section \ref{CAR}).

\subsubsection{Self Training (ST)}
Self-training has not a clear definition is the literature, it can be viewed
as a “single-view weakly supervised algorithm”. First a classifier is trained  from  the available labeled examples and then this classifier is used to make predictions and build new proxy labels.
%Only those instances with a labeling confidence exceeding a certain threshold are added to the labeled set.
%Only the examples with a proxy-labeling confidence exceeding a certain threshold are added to the training set.
Only those examples where confidence in proxy labelling exceeds a certain threshold are added to the training set.
Then, the classifier is retrained from the training set enriched with proxy-labels. 
This process is repeated in an
iterative way \cite{ennaji2012self}.  %could continues for several iterations. 
% ?? utile ?? The classifier could be simply a one nearest neighbor as in \cite{ennaji2012self}.

\subsubsection{Co-Training (CT) \cite{blum1998combining,davy2005review,ZHAO2017,Mihalcea2004}}
%Starting with a set of labeled data, co-training algorithms attempt to increase the amount of labeled data using some amounts of unlabeled data.
Starting from a set of partially labeled examples, co-learning algorithms aim to increase the amount of labeled examples by generating proxy-labels. 

%\begin{figure}[h]
%\centering
%\includegraphics[width=0.8\linewidth]{./Figures/cotraining.png}
%\caption{Illustration of cotraining}
%\label{cotraining}
%\end{figure}

%\begin{table*}[!]
%\centering
%{\small
%\fontsize{9}{10}\selectfont
%\begin{tabular}{|l|c|c|c|c|c|c|c|c|c|c|c|}
%\hline
%                                        & Mentioned  &  &  &  &  &  &  &  &  &   & ... \\
%                                        & in Sections & MIL & RLL & TL & AL & SSL & ST & CT & PUL &  MIX & ... \\\hline
%Inaccurate labels (“bag of examples”)   &&  X  &     &    &    &     &    &    &     &   & \\\hline
%Inaccurate labels ("noisy")             &&     & X   &    &    &     &    &    &     &   & \\\hline
%Inaccurate labels ("partial-label learning") &&     & X   &    &    &   X  &    &    &     &   & \\\hline
%Inexact labeling ("proxy domain")       &&     &     &  X &    &     &    &    &     &   & \\\hline
%Inexact labeling ("proxy labels")       &&     &     &    &    &     &  X &  X &     & X & \\\hline
%Inexact labeling ("proxy individuals")  &&  X  &     &    &    &     &    &    &     & X & \\\hline
%Incomplete number of labels             &&     &     &    &  X &  X  &  X &  X &  X  & X & \\\hline
%...                                     &&     &     &    &    &     &    &    &     &   & \\\hline
%\end{tabular}
%\caption{Main type of weakness versus its main useful learning algorithm.}}
%\label{weaknessvslearning}
%\end{table*}

%Co-training algorithms work by generating several classifiers trained on the input labeled data, which are then used to label new unlabeled data. 
Co-training algorithms work by training several classifiers from the initial labeled examples. Then, these classifier are used to make predictions and generate proxy-labels on the unlabeled examples. The most confident predictions on these proxy-labels are then added to the set of labeled data for the next iteration.
%From this new labeled data, the most confident predictions are kept, and subsequently added to the set of labeled data. 
%The most confident predictions of a fixed classifier are used to generate labels which are used by the other classifiers. \textcolor{blue}{AO: je n'ai pas compris cette derniere phrase. 2 "used"}
%This iterative process is illustrated in Figure \ref{cotraining}.

One important aspect of co-training is the relationship between the views (the sets of explicative variables) used in learning the different models. 
%In the original definition of the initial co-training algorithm \cite{blum1998combining}  state conditional independence of the views as a required criterion for co-training to work. 
The original co-training algorithm \cite{blum1998combining}  states that the independence of the views is required to
properly perform automatic labeling.
More recent works \cite{nigam2000analyzing,Abney02,Clark03} show that this assumption can be relaxed.
Another requirement is to obtain at the iteration step a ``reasonable'' classifier in terms of performances , this explains why we place
co-training on the left of AL and SSL in Figure~\ref{quantity}.
In \cite{ng-cardie-2003-weakly}, a study is given on the optimal selection of the co-training parameters.

Co-training can also be considered as a member of "multi-view training" family in which some other members belong to, such as: Democratic Co-learning \cite{Zhou2004Democratic}, Tri-training \cite{Zhou2005TriTraining}, Asymmetric tri-training \cite{saito2017asymmetric}, Multi-task tri-training \cite{ruder-plank-2018-strong}, which are not described here.

% il y a 3 paramètres à régler dans le cotraining en parle-t-on ? voir \cite{Mihalcea2004}

\section{Other key elements - Beyond the 3 axes}

%\textcolor{magenta}{small text to add}

\subsection{Learning at the crossroad of the three axis}

%The use of a cube as a descriptive figure for Weakly Supervised Learning allows one to not only use axis to outline or explore the literature but the space as a whole. Indeed when some sub-fields or ideas lie in between two axis, we can just place them in the plan described by these two. For example Partial Label Learning could be related to two supervision deficiencies. The first one being inexact supervision as multiple label are provided for every samples and not only the true label. The second one being inaccurate supervision as only one of all provided labels is the correct one. Having the whole plan defined by these two latter axes seems more convenient to describe PLL.
%
The use of a cube to describe the literature on Weakly Supervised Learning allows us not only to use the axes, but also the volume of the cube to position existing approaches. 
%This allows us to more subtly position approaches that are related to several axes at once. 
It is now easy to position more subtly the approaches that are related to several axes at once.
For example, Partial Label Learning may be related to two supervision deficiencies: i) inexact supervision, because multiple labels are provided for each training example; ii) inaccurate supervision, because only one of the labels provided is correct. Positioning the PLL on the plane defined by these two axes seems more relevant. 
%

%This representation also helps to highlight some points or axis when they are defined by intersections, between two axes or between an axis and a plan. One of these key points is the origin of all Axes, this point correspond to the case where the supervision is totally inaccurate, inexact and incomplete which is unsupervised learning. A similar point stands out at the infinite end of the cube, representing a totally accurate, exact and complete supervision for supervised learning.

Also, this representation allows to highlight some interesting intersections, between two axes or between an axis and a plane. One of these points of interest is the origin of the three axes, which corresponds to the case where supervision is absolutely inaccurate, imprecise and incomplete, which ultimately amounts to unsupervised learning. Similarly, the point at the opposite end of the cube corresponds to perfectly precise, accurate and complete supervision, which equates to supervised learning.

%RQ Alexis : faire apparaitre ces points sur la figure ? le petit paragraphe qui les positionne n'est pas si clair .. 

%Finally it could provide insights on why using proven techniques from a particular subfield of Weakly Supervised Learning can be efficient in another one. DivideMix \cite{li2020dividemix}, for example, chose to reuse the efficient MixUp \cite{zhang2018mixup} approach from Semi-Supervised Learning to Learning with Label Noise. This approach use Data Augmentation \cite{shorten2019survey} and Model Agreement \cite{qiao2018deep} to estimate label probability and then discard or keep provided labels. Finally it either use them as regularization for unlabelled data or in the supervision task for labelled data.

Finally this representation could provide insights on the reasons of
using proven techniques from a particular subfield of Weakly Supervised Learning can be efficient in another one. For instance, DivideMix \cite{li2020dividemix} chooses to reuse the efficient MixUp \cite{zhang2018mixup} approach from Semi-Supervised Learning to tackle the problem of Learning with Label Noise. This approach uses Data Augmentation \cite{shorten2019survey} and Model Agreement \cite{qiao2018deep} to estimate labels probabilities and then discard or keep provided labels. 

%Note that due to page limit, this section actually serves more like an example of points inside the bube rather than a  comprehensive  review.   Readers  interested are encouraged to read the corresponding  and  other references.
This section is not exhaustive, interested readers will be able to position the approaches of the literature in the cube themselves.

%RQ Alexis : ces deux exemples, DivideMix et PLL sont bien, mais il faudrait en donner d'avantage d'intuitions ... ce qui est écrit pour le moment est un peu sec / court .. 

%RQ Alexis : très bien l'argument de dire que les différentes approches de WSL ont une certaine "porosité" , et ce qu'on propose peut permettre de comprendre les interaction entre les sous-domaine du WSL :-) Very nice :-)

%RQ Vincent: insister que ce ne sont que des examples et que le lecteur puisse en imaginer d'autres

\subsection{Deficiency Model}
\label{CAR}

The deficiency model describes the nature %of the origin 
of the supervision deficiency. It is usually described as a probability measure called $\rho : (x,y) \mapsto \rho(x,y)$, indicating if an example is corrupt or not. 
$\rho$ can depends on the value the explanatory variables $x \in \mathcal{X}$, 
the label value $y \in \mathcal{Y}$ or both $(x,y)$. 
The different types of supervision deficiency described in this section are the following: (i) Completely At Random (CAR), (ii) At Random (AR) and (iii) Not At Random (NAR).

If the probability of being corrupted is the same for all training examples, $\rho : (x,y) \mapsto \rho_c$, $\rho_c \in [0,1]$, then the supervision deficiency model is Completely At Random (CAR). This implies that the cause of the supervision deficiency data is unrelated to the data. If the probability of being corrupted is the same within classes, $\rho : (x,y) \mapsto \rho_y$, $\forall y \in \mathcal{Y}$, $\rho_y \in [0,1]$, then the supervision deficiency model is At Random (AR). If neither CAR nor AR holds, then we speak of Not at Random (NAR) model. Here the probability of being corrupted is dependent on both the samples and the label value, $\rho: (x,y) \mapsto \rho(x,y)$.
These three deficiency models can be ranked in a descendent manner, having the NAR model being the most complex as it depends on both the instance and label value, which requires a function to model, to CAR model where only one constant is enough to describe it.
These models may help practitioner to find links between supervision deficiencies. For example PUL is SSL with only one class labeled, which means that the missingness of the label is linked to the label value, so PUL is an extreme case of SSL AR with $\rho_0=1-e$ and $\rho_1=e$ (where $e$ is called the propensity score).
%When $\rho_0=1-e$ and $\rho_1=e$ (where $e$ is called the propensity score)  PUL is an extreme case of SSL AR.

AL is another form of SSL where examples are labeled thanks to a strategy, previously labeled instances and the ordered iterative process leading to non-iid labeled data. As such AL is part of the SSL NAR family.
We want to reiterate the deficiency model can be applied to any supervision deficiency, even if it has been mostly featured in RLL and in SSL.

% RQ Antoine : D'abord les présenter puis dire qu'ils sont de compléxité ascendante.

% RQ Antoine : Changer les notations du rho

%RQ Alexis : ne pas oposer les deux : le cube et le type de "supervision deficiency" .... plutôt dire, ce qui suit est applicable dans tout le cube ... 

% RQ Alexis : très bien :-) préciser que PUL est un cas extrême ? 

% RQ Alexis : bon exemple, mais a expliquer d'avantage car je ne suis pas certain que le lecteur lambda va comprendre ... 

% RQ Alexis : !!!! ça fil faudrait le dire dès le début ?? 

% RQ Alexis : très bien ! peu etre dire au début du paragaphe que ces trois types de corruption sont présentés dans un ordre croissant de difficulté ? Et peut être exprimer ces trois idée en introduisant des notation, ex : exprimer la proba d'être corrompu , = une constante pour le CAR, = p(.|y) pour le AR, et = p(.|x,y) pour le NAR ?? 
%
% RQ Antoine : D'abord les présenter puis dire qu'ils sont de compléxité ascendante.

\subsection{Transductive learning vs. Inductive Learning}

% RQ Alexis : je me demande si cette section n'est pas un peu longue car cette notion est importante mais assez simple à comprendre ? 

%Since we are in this paper in the Weakly Supervised Learning 
As we consider WSL framework, one may be tempted to use the test set to guide the choice of the model. But in this case we need to carefully decide if in the future the need of a model to predict on another test (deployment) dataset is required or not: two point of view could be considered transductive learning vs. inductive learning, that is why now we add a note on them.

Training a machine could take many forms as supervised learning, unsupervised learning, active learning, online learning, etc.  The number of members is the family is large and new members appear regularly as for example “federative learning”. However one may establish a separation between two constant classes based on the way the user would like to use the “learning machine” at the deployment stage. The user does not want necessarily a predictive model for subsequent use on new data. Because, for example, it has the completeness of the data for the problem to be treated. It is therefore necessary to distinguish between inductive learning and transductive learning.

On one side the goal of inductive learning is, essentially, to learn a function (a model) which will be later used on new data to predict classes (classification) or numerical values (regression). The predictions may be seen as “buy-products” of the model. Induction is reasoning from observed training cases to general rules, which are then applied to the test cases.  On the
other side the goal of transductive learning the goal is not to obtain a function or a model but only to do predictions on a given test database, and on only on this test of instances. Transduction was introduced by Vladimir Vapnik in the 1990s, motivated by the intuition 
that transduction is preferable to induction since, %according to him, 
induction requires solving a more general problem (inferring a function) before solving a more specific problem (computing outputs for new cases). However the distinction between inductive and transductive learning could be a hazy border for example in case of semi-supervised learning. Knowing this, the view of Zhou in \cite{zhou2017} about ``pure semi supervised learning'' and transductive learning is interesting.
The distinction about Transductive learning vs. Inductive Learning concerned most of the learning form included on Figure~\ref{bigpicture}.

%%%%%%%%%%%%%%%
%\subsection{Resume}

%%% TODO : tout les croisement dans le cube sont possibles ... ajouter des examples

%%%%%%%%%%%%%%%%%%%%%%%%%%%%%%%%%%%%%%%%%%%%%%%%%%%%%%%%%%%%%%%%%%%%%%%%%%%%%%%%%%%%%%%%%%%%%%%%%%%%%%%%%%%%%%%%%%%%%%%%%%%%%%%%%%%%%%%%%%%%%%%%%%%%%%%%%%%%%%%%%%%%%%%%%%%%%%%%%%%%%%%%%%%%%%%%%%%%%%%%%%%%%%%%%%%%%%%%%%%%%%%

\section{The 3 common concepts of WSL}
\label{3concepts}

Until now we see that many forms
of learning and weakness are intertwined. A way to resume their aspect was given on Figure~\ref{bigpicture}.
%It also possible to try to give other way of view of these three axis
%in a table (see Table \ref{weaknessvslearning}).
%As this table shows different form of learning may address the same weakness and the opposite and some recoveries exist.
From this point of view one may identified 3 common concepts that are described now.

\subsection{Quantity $|L|$}
\label{quantity}

Insufficient quantity of labels or training examples occurs when many training examples are available but only a small portion is labeled, e.g. due to the cost of labelling. For instance, this occurs in the field of cyber security where human forensics is needed to tag attacks. Usually, this issue is addressed by few shot learning (FSL), active learning (AL) \cite{settles2009active} semi-supervised learning (SSL) \cite{chapelle2009semi} , Self Training, or Co-Training or active learning (AL) which have been described briefly above in this paper. Another way to see the "quantity"
could be the ratio between the number of examples labeled and unlabeled ($p$).

%\textcolor{blue}{AO: les parametres $q,a et p$ sont definis apres leurs usages, il faut indiquer leurs references}
%\textcolor{green}{PN : il me semble pas}

\subsection{Quality $q$}
\label{quality}

In this case, all {training examples} are labeled but the labels may be corrupted. %For example, 
This usually happens when outsourcing labeling to crowd labeling \cite{pmlr-v22-urner12}. The Robust Learning to Label Noise (RLL) approaches tackle this problem \cite{frenay2013classification}, with three types of label noise identified: i) the \textit{completely at random} noise corresponds to a uniform probability of label change ; ii) the \textit{class dependent} label noise %is 
when the probability of label change depends upon each class, with uniform label changes within each class ; iii) the \textit{instance dependent} label noise is when the probability of label change varies over the input space of the classifier.  This last type of label noise is the most difficult to deal with, and typically requires  making sometimes strong assumptions on the data.

\subsection{Adaptability $a$}
This is the case for instance, in Multi Instance Learning (MIL) \cite{yang2005review,zhou2006multi,foulds2010review,Carbonneau_2018}, in which there is one label for each bag of training examples, and each example has an uncertain label.
Some scenarios in  Transfer Learning (TL) \cite{weiss2016survey} imply that only the labels in the source domain are provided while the target domain labels are not. Often, these non-adapted labels are associated with the existence of slightly different learning tasks \textit{(e.g. more precise and numerous classes are dividing the original categories)}. 
Alternatively, non-adapted labels may characterize a differing statistical individual \cite{Conte2010} \textit{(e.g. a subpart of an image instead of the entire image)}. 

%%%%%%%%%%%%%%%%%%%%%%%%%%%%%%%%%%%%%%%%%%%%%%%%%%%%%%%%%%%%%%%%%%%%%%%%%%%%%%%%%%%%%%%%%%%%%%%%%%%%%%%%%%%%%%%%%%%%%%%%%%%%%%%%%%%%%%%%%%%%%%%%%%%%%%%%%%%%%%%%%%%%%%%%%%%%%%%%%%%%%%%%%%%%%%%%%%%%%%%%%%%%%%%%%%%%%%%%%%%%%%%
\section{From WSL to Biquality learning (when $a=1$)}
All the types of supervision deficiencies presented above %in this paper 
are addressed separately in the literature, leading to highly specialized approaches. 
In practice, it is very difficult to identify the type(s) of deficiencies with which a real dataset is associated. 
For this reason, it would be very useful to suggest another point of view as a tentative of an unified framework for (a part of the) Weakly Supervised Learning, in order to design generic approaches capable of dealing not a single type of supervision deficiency. This is the purpose of this section mainly given for cases where data are adapted to the task to learn ($a=1$).

Learning using biquality data has recently been put forward in \cite{Charikar2016,Hendrycks2018,Hataya2019UnifyingSA} and consists in learning a classifier from two distinct training sets, one trusted and the other not.
The initial motivation
 was to unify semi-supervised and robust learning through a combination  of the two.  We show in this paper that this scenario is not limited to this unification and that it can cover a larger range of supervision deficiencies as demonstrated with the algorithms we suggest and their results. 

The trusted dataset $D_T$ consists of pairs of labeled examples ($x_i, y_i$) where all labels $y_i \in \mathcal{Y}$ are supposed to be correct according to the true underlying conditional distribution $\mathbb{P}_T(Y|X)$. In the untrusted dataset $D_U$, examples $x_i$ may be associated with incorrect labels. We note $\mathbb{P}_U(Y|X)$ the corresponding conditional distribution.

At this stage, no assumption is made about the nature of the supervision deficiencies which could be of any type including label noise, missing labels, concept drift, non-adapted labels  ... and more generally a mixture of these supervision deficiencies.

The difficulty of a learning task performed on biquality data can be characterised by two quantities.
First, the ratio of trusted data over the whole data set, denoted by $p$: 
\begin{equation}
\small{
p = \frac {|D_T|}{|D_T|+|D_U|}
}
\end{equation}

Second, a measure of the quality, denoted by $q$, which evaluates the usefulness of the untrusted data $D_U$ to learn the trusted concept For example in \cite{Hataya2019UnifyingSA} $q$ is defined using a ratio of Kullback-Leibler divergence between $\mathbb{P}_T(Y|X)$ and $\mathbb{P}_U(Y|X)$.

\begin{figure}[!h]
\centering
\includegraphics[width=0.85\linewidth]{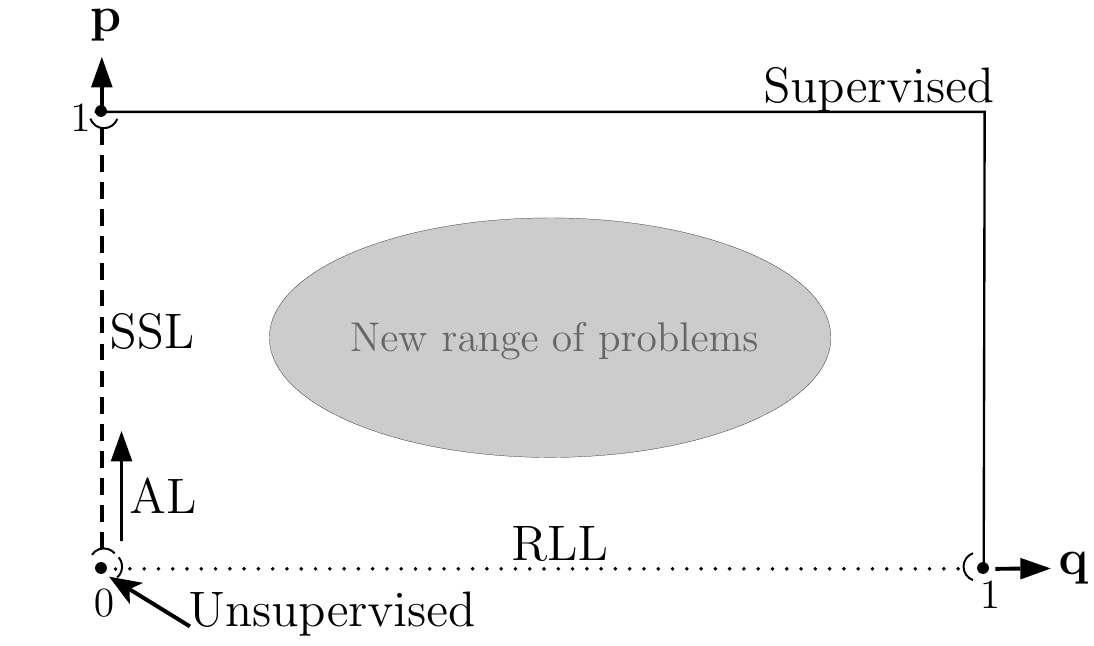}
\caption{{The different learning tasks covered by the biquality setting, represented on a 2D representation.}}
\label{2D}
\end{figure}

The biquality setting covers a wide range of learning tasks by varying the quantities $q$ and $p$, as represented in Figure \ref{2D}.  

\begin{itemize}
  \item When ($p=1$ OR $q=1$)  all examples can be trusted. %Thus, 
  This setting corresponds to a standard {\bf supervised learning} (SL) task.
  
  \item When ($p=0$ AND $q=0$), there is no trusted {examples} and the untrusted labels are not informative. We are left with only the inputs $\{x_i\}_{1 \leq i \leq m}$ as in {\bf unsupervised learning (UL)}.
  
  \item On {the vertical axis defined by} $q=0$, except for the two points $(p,q)=(0,0)$ and $(p,q)=(1,0)$, the untrusted labels are not informative, and trusted examples are available. The learning task becomes {\bf semi-supervised learning (SSL)} with the untrusted examples as unlabeled and the trusted as labeled.
  
  \item An upward move on this vertical axis, from a point $(p,q)=(\epsilon,0)$ characterized by a low proportion of labeled examples $p=\epsilon$, to a point $(p',0)$, with $p' > p$, corresponds to {\bf Active Learning}, if an oracle can be called on unlabeled  examples. The same upward move can also be realized in {\bf Self-training} and {\bf Co-training}, where unlabeled training examples are labeled using the predictions of the current classifier(s). %{RQ Alex : le début de du paragraphe AL ne me parait si clair. (première phrase un peu longue ?)} 
  
  \item On the horizontal axis {defined by} $p=0$, {except for the points} $(p,q)=(0,0)$ and $(p,q)=(0,1)$, only untrusted examples are provided, which corresponds to the range of learning tasks typically addressed by {\bf Robust Learning to Label noise (RLL)} approaches. 
  \end{itemize}

Only the edges of Figure \ref{2D} have been envisioned in previous works {--  i.e.} the points mentioned above {--} and a whole new range of problems corresponding to the entire plan of the figure remains to be explored.
Biquality {learning} may also be used {to tackle particular tasks belonging to WSL}, for instance:

\begin{itemize}
  \item Positive Unlabeled Learning (PUL) \cite{Bekker_2020} where the trusted examples are only positive and untrusted examples those from the unknown class. 
    
  \item Self Training and Co-training \cite{blum1998combining,davy2005review,ZHAO2017} could be adressed at the end of their self labeled process: the initial training set is the trusted dataset, all examples labeled after (during the self labeling process) are the untrusted examples. %\todo{RQ Antoine : déjà cité avant)}
  
  \item Concept drift \cite{Gama2014}: when a concept drift occurs, all the examples used before a drift detection may be considered as the untrusted examples, while the examples available after it are viewed as the trusted ones, assuming a perfect labeling process. %Note that close to the instant detection $p$ is very low and grows up as time pass. 
  
  \item Self Supervised Learning system as Snorkel \cite{ratner2020snorkel}: the small initial training set is the trusted dataset, all examples automatically labeled using the labeling functions correspond to the untrusted examples.
  %\item Noisy Labels ? 
  %\todo{(RQ Antoine : déjà cité avant)}
\end{itemize}

As can be seen from the above list, the Biquality framework is quite general and its investigation seems a promising avenue to unify different aspects of the  Weakly Supervised Learning.

%%%%%%%%%%%%%%%%%%%%%%%%%%%%%%%%%%%%%%%%%%%%%%%%%%%%%%%%%%%%%%%%%%%%%%%%%%%%%%%%%%%%%%%%%%%%%%%%%%%%%%%%%%%%%%%%%%%%%%%%%%%%%%%%%%%%%%%%%%%%%%%%%%%%%%%%%%%%%%%%%%%%%%%%%%%%%%%%%%%%%%%%%%%%%%%%%%%%%%%%%%%%%%%%%%%%%%%%%%%%%%%
\section{Biquality Learning - Existing Works}

In the previous section we have been describing how Wealky Supervised Learning subfields fitted in the Biquality Learning setup. Here we would be reviewing three of these subfields and highlight prexisting Biquality Learning algorithms that either have been made for a different purpose but still could be used for WSL, or have been design directly for this setup.

%, pour chaque sous section :
% donner l'intuition générale du lien avec le biquality learning en "high level", puis un exemple d'article, de méthode ou d'algo sans trop de détail?

% 

\subsection{Transfer Learning}

% pour cette section je mettrai donc :
% 1) intuition générale du lien avec le biquality learning en "high level"
% 2) qui inclurait trois bullets
% transfert learning ?
% multi task learning ?
% fine tuning ?
%
% le petit paragraphe sur tradaboost me parrait dès lors inutile mais la ref reste
% 

Transfer Learning focuses on storing, knowledge gained while solving one problem and applying it to a different but related problem. Two datasets are at disposal, a source dataset $D_S$ and target dataset $D_T$ that are related to a source domain $\mathcal{D}_S(\mathcal{X}_S,\mathbb{P}(X_S))$ and a target domain $\mathcal{D}_T(\mathcal{X}_T,\mathbb{P}(X_T))$ to solve the target task $\mathcal{T}_T(\mathcal{Y}_T,\mathbb{P}(Y_T|X_T))$ with the help of the source task $\mathcal{T}_S(\mathcal{Y}_S,\mathbb{P}(Y_S|X_S))$.
We can draw a parallel between Biquality Learning notations and Transfer Learning notations mostly by substituting (source, $S$) by (untrusted, $U$) and (target, $T$) by (trusted, $T$).

A lot of different setups can derive from the general Transfer Learning setup as Domain Adaptation, Transductive Transfer Learning, Covariate Shift, ... Inductive Transfer Learning is the setup closest to Biquality Learning, indeed most of the key assumptions are the same : $\mathcal{X}_T=\mathcal{X}_U$, $\mathcal{Y}_T=\mathcal{Y}_U$, $\mathbb{P}(X_T)=\mathbb{P}(X_U)$, $\mathbb{P}(Y_T|X_T)\neq\mathbb{P}(Y_U|X_U)$.

For example, TrAdaBoost \cite{dai2007boosting} is an extension of boosting to Inductive Transfer Learning. TrAdaBoost learns on both trusted and untrusted data every iterations. It behaves exactly as AdaBoost \cite{freund1997decision,hastie2009multi} on trusted data : mispredicted trusted samples get more attention, but opposite on untrusted data : mispredicted untrusted samples are ditched out.

Multi Task Learning \cite{caruana1997multitask} is another Inductive Transfer Learning approach that improves generalization by learning both tasks in parallel while using a shared representation; what is learned for the untrusted task can help the trusted task. This loss $L_{\text{MTL}}$ is usually defined by a convex combination of the trusted loss $L_T$ and untrusted loss $L_U$ of the model $f$ (with $0\le\lambda\le 1$):
\begin{equation}
\medmath{
    L_{\text{MTL}}(f(X),Y) = (1-\lambda)L_U(f(X),Y) + \lambda L_T(f(X),Y)
}
\end{equation}

In Inductive Transfer Learning as in Transfer Learning in general, we assume that the source task (i.e. untrusted task) is relevant for the target task (i.e. trusted task). Nonetheless in the Biquality Data setup, we can have the untrusted task that bring no information to the trusted task, even bring adversarial information. Thus using Inductive Transfer Learning algorithm directly on Biquality Data setup can lead to bad predictive performances.

For example, with Multi Task Learning, the global loss term would be heavily perturbed as the untrusted loss could never be optimized. For TrAdaBoost, the first model learned on both trusted and untrusted samples would not be able to learn the class boundaries correctly, and the weight updating schemes would not be efficient.

\subsection{RLL and Transition Matrix}

% RQ vincent
% pour cette section je mettrai donc :
% 1) intuition générale du lien avec le biquality learning en "high level"
% pas trop de chose à retirer

A family of Biquality Learning algorithm has been pioneered by Patrini with \cite{patrini2017making} from the Robust Learning to Label Noise literature. These algorithms try to estimate the per class probabilities of label flip into another class (of the $K$ classes) which defines the Transition Matrix $T$.
\begin{equation}
\medmath{
    \forall (i,j) \in K^2,\: T_{(i,j)} = \mathbb{P}(Y_U=j|Y_T=i)
\label{transition-matrix-definition}
}
\end{equation}

Patrini proposed in \cite{patrini2017making} to used the Transition Matrix $T$ to adapt any supervised loss functions $L$ to learning with label noise. The two corrections proposed are : (i) the forward loss correction: {\small $L^{\rightarrow}(f(X),Y)=L(T^{\top}\cdot f(X),Y)$ } and (ii) the backward loss correction: {\small $L^{\leftarrow}(f(X),Y) = T^{-1}\cdot L(f(X),Y)$}.

When no trusted samples are available as in \cite{patrini2017making}, Patrini proposed to use anchor points in order to estimate $T$. An anchor point from the $i$-th class is the point with the highest probability to be from the $i$-th class from a given dataset.
\begin{equation}
\medmath{
    \forall i \in K,\: A_i = \operatorname*{argmax}_x \mathbb{P}(Y=i|X=x)
\label{anchor-points-definition}
}
\end{equation}

Thanks to this definition Patrini propose an estimator of the Transition Matrix : 
\begin{equation}
\medmath{
    \hat{T}_{(i,j)} = \mathbb{P}(Y=j|X=A_i)
    \label{transition-matrix-patrini-estimator}
}
\end{equation}

Finally the procedure to learn a model $f$ that minimizes $L$ on untrusted data with Patrini's approach is in two steps. First learn $f$ model on untrusted data with a loss $L$. Estimate the Transition Matrix thanks to $\hat{T}$ with Equation \ref{transition-matrix-patrini-estimator}. Then learn a model $f$ with either $L^{\rightarrow}$ or $L^{\leftarrow}$.

This algorithm, designed for Robust Learning to Label noise can easily be adapted to Biquality Learning. Hendrycks proposed one adaptation in \cite{Hendrycks2018} with some changes to Patrini's approach.
As trusted data are available, there is no more the need to use anchor points to represent our trusted concept. So another estimator for the Transition Matrix is proposed by learning a model $f_U$ on untrusted data, and making probabilistic predictions with $f_U$ on the trusted dataset $D_T$ and comparing it to the trusted labels $y_T$ :

\begin{equation}
\medmath{
\hat{T}_{(i,*)} = \sum_{x_i \in D^i_T}\frac{f_U(x_i)}{\sum_{z_i \in D^i_T}||f_U(z_i)||}
}
\label{transition-matrix-hendrycks-estimator}
\end{equation}
where $D^i_T = \{\forall (x,y) \in D_T | y=i\}$.
Then for the final step, Hendrycks proposed to learn $f$ with the corrected forward loss $L^{\rightarrow}$ on the untrusted data, and the uncorrected loss $L$ on the trusted data.
Thus GLC is an example of a Biquality Learning algorithm that has been demonstrated to be quite efficient on At Random supervision deficiencies.

\subsection{Covariate Shift}

% RQ vincent
% pour cette section je mettrai donc :
% 1) intuition générale du lien avec le biquality learning en "high level"
% trop de détail de mon point de vue sur IRBL
% IRBL covariate shift Or concept drift ?

Covariate Shift literature has also inspired people to adapt these algorithms to Biquality Learning. The algorithm with the most influence in this regard is called Importance Reweighting \cite{4803844}, which aims was to give %a 
high weights to source samples that were similar to the target samples, and low weights when they were not similar. This objective fits well with Not At Random (or sample dependent) corruptions as the correction made to the untrusted dataset is per sample with this algorithm family. Multiple approaches has been inspired by this literature. 

The key idea of this algorithm family is to define a loss function $\tilde{L}$ such that learning a model $f$ on $D_U$ that minimizes $\tilde{L}$ is equivalent to using the original loss function $L$ on $D_T$ in the risk estimate. The following equations show how $\tilde{L}$ appears from the risk estimate $R$:  

\begin{equation}
\label{reweighting-risk-equation}
\medmath{
\begin{aligned}
&R_{(X,Y)\sim T,L}(f) = \mathbb{E}_{(X,Y)\sim T}[L(f(X),Y)]\\
&\qquad = \mathbb{E}_{(X,Y)\sim U}[\frac{\mathbb{P}_T(X,Y)}{\mathbb{P}_U(X,Y)}L(f(X),Y)]\\
&\qquad = \mathbb{E}_{(X,Y)\sim U}[\beta L(f(X),Y)]\\
&\qquad = R_{(X,Y)\sim U,\tilde{L}}(f)\\
\end{aligned}
}
\end{equation}

However this newly defined loss function $\tilde{L}$ can be hard to estimate and thus approaches have been proposed to further simplify the weight estimation.

For example, Importance Reweighting for Biquality Learning (IRBL) \cite{nodet2021importance} uses the biquality hypothesis that the distribution $\mathbb{P}(X)$ is the same in the trusted and untrusted datasets. By using Bayes Formula we have a new expression for $\beta$ :

%\begin{equation}
%\label{irbl-equation}
%{\small
%\begin{aligned}
%\beta_{\text{IRBL}}&=\frac{\mathbb{P}_T(X,Y)}{\mathbb{P}_U(X,Y)}\\
%&=\frac{\mathbb{P}_T(Y|X)\mathbb{P}(X)}{\mathbb{P}_U(Y|X)\mathbb{P}(X)}\\
%&=\frac{\mathbb{P}_T(Y|X)}{\mathbb{P}_U(Y|X)}\\
%\end{aligned}
%}
%\end{equation}

\begin{equation}
\label{irbl-equation}
\medmath{
\beta_{\text{IRBL}}=\frac{\mathbb{P}_T(X,Y)}{\mathbb{P}_U(X,Y)}=\frac{\mathbb{P}_T(Y|X)\mathbb{P}(X)}{\mathbb{P}_U(Y|X)\mathbb{P}(X)}=\frac{\mathbb{P}_T(Y|X)}{\mathbb{P}_U(Y|X)}
}
\end{equation}

First, the vector of ratios between $\mathbb{P}_T(Y|X)$ and $\mathbb{P}_U(Y|X)$ is estimated by the term ${\small f_T(x_i)\oslash f_U(x_i)}$, using the models $f_T$ and $f_U$ learn on $D_T$ and $D_U$. For each untrusted example, the weight $\hat{\beta}_{\text{IRBL}}$ is the $y_i$-th element of this vector; while  $\hat{\beta}_{\text{IRBL}}$ is fixed to 1 for the trusted examples. Then, the final classifier is learned from $D_T \cup D_U$ by minimizing $\tilde{L}$.

Another algorithm has been proposed in \cite{fang2020rethinking} named Dynamic Importance Reweighting (DIW) by writing Equation \ref{irbl-equation} in a more traditional way with Bayes Formula.

%\begin{equation}
%\label{diw-equation}
%{\small
%\begin{aligned}
%\beta_{\text{DIW}}&=\frac{\mathbb{P}_T(X,Y)}{\mathbb{P}_U(X,Y)}\\
%&=\frac{\mathbb{P}_T(X|Y)\mathbb{P}_T(X)}{\mathbb{P}_U(X|Y)\mathbb{P}_U(Y)}\\
%\end{aligned}
%}
%end{equation}

\begin{equation}
\label{diw-equation}
\medmath{
\beta_{\text{DIW}}=\frac{\mathbb{P}_T(X,Y)}{\mathbb{P}_U(X,Y)}
 \; = \; \frac{\mathbb{P}_T(X|Y)\mathbb{P}_T(Y)}{\mathbb{P}_U(X|Y)\mathbb{P}_U(Y)}}
%\end{medmath}
\end{equation}

To estimate $\beta_{\text{DIW}}$, the trick is to select both sub-samples of $D_T$ and $D_U$ with samples of the same classes and then use an Density Ratio Estimator \cite{sugiyama2010density} such as Kernel Mean Matching (KMM) \cite{huang2007correcting,gretton2009covariate}. Then a final classifier is learned on $D_U$ by minimizing $\tilde{L}$.
One particular issue of this algorithm is that KMM is learned by optimizing a quadratic program, $K$-times per batch, that leads to high algorithm complexity especially in the case of massive multiclass classification.

IRBL and DIW are two new Biquality Learning algorithms that work on NAR cases.

%\newpage
%%%%%%%%%%%%%%%%%%%%%%%%%%%%%%%%%%%%%%%%%%%%%%%%%%%%%%%%%%%%%%%%%%%%%%%%%%%%%%%%%%%%%%%%%%%%%%%%%%%%%%%%%%%%%%%%%%%%%%%%%%%%%%%%%%%%%%%%%%%%%%%%%%%%%%%%%%%%%%%%%%%%%%%%%%%%%%%%%%%%%%%%%%%%%%%%%%%%%%%%%%%%%%%%%%%%%%%%%%%%%%%
%\section{Suggestion of a generic framework}

%One contribution of this paper is to suggest a generic framework for achieving biquality learning and thus covering many facets of WSL. This is presented in this section.

%reprendre AAAI ?

%%%%%%%%%%%%%%%%%%%%%%%%%%%%%%%%%%%%%%%%%%%%%%%%%%%%%%%%%%%%%%%%%%%%%%%%%%%%%%%%%%%%%%%%%%%%%%%%%%%%%%%%%%%%%%%%%%%%%%%%%%%%%%%%%%%%%%%%%%%%%%%%%%%%%%%%%%%%%%%%%%%%%%%%%%%%%%%%%%%%%%%%%%%%%%%%%%%%%%%%%%%%%%%%%%%%%%%%%%%%%%%
\section{Concluding remarks}

%\textcolor{blue}{
%In this paper, we discussed ...
% we gave an overview of WSL and its links to Biquality Learning
In this paper, we propose a unified view of Weak Supervised Learning to cope with the shortcomings of the supervision in the field of Machine Learning. We discussed these shortcomings through a cube
%\emph{cube} 
along with three axes corresponding to the characteristics of training
%\it{(pas certain de moi le training, learning à la place?)}
labels (inaccurate, inexact and
%\it{(ou “and”?)} 
incomplete). The detailed presentation of these axes gives an insight the different existing learning approaches which can be more subtly position on the cube. In this way, the links between some subfields of WSL with Biquality Learning are highlighted, showing how the algorithms of the latter field can be used within the framework of WSL.
%}

%\textcolor{orange}{Together - }
%For biquality learning, we call the WSL  community to consider the following action points for research:
%
%%    \item 
%\end{itemize}

% parler de la commercialisation ? point qui montre que c'est pas facile ?
%
%
%MAX 10 Pages avec les reférences !

\bibliographystyle{IEEEtran}
\bibliography{references}

\begin{thebibliography}{10}
\providecommand{\url}[1]{#1}
\csname url@samestyle\endcsname
\providecommand{\newblock}{\relax}
\providecommand{\bibinfo}[2]{#2}
\providecommand{\BIBentrySTDinterwordspacing}{\spaceskip=0pt\relax}
\providecommand{\BIBentryALTinterwordstretchfactor}{4}
\providecommand{\BIBentryALTinterwordspacing}{\spaceskip=\fontdimen2\font plus
\BIBentryALTinterwordstretchfactor\fontdimen3\font minus
  \fontdimen4\font\relax}
\providecommand{\BIBforeignlanguage}[2]{{%
\expandafter\ifx\csname l@#1\endcsname\relax
\typeout{** WARNING: IEEEtran.bst: No hyphenation pattern has been}%
\typeout{** loaded for the language `#1'. Using the pattern for}%
\typeout{** the default language instead.}%
\else
\language=\csname l@#1\endcsname
\fi
#2}}
\providecommand{\BIBdecl}{\relax}
\BIBdecl

\bibitem{SugiyamaTalkIdiap}
\BIBentryALTinterwordspacing
M.~Sugiyama, ``Talk: Recent advances in weakly-supervised learning and reliable
  learning,'' 2019. [Online]. Available:
  \url{https://portal.klewel.com/watch/webcast/recent-advances-in-weakly-supervised-learning-and-reliable-learning/talk/1/}
\BIBentrySTDinterwordspacing

\bibitem{zhou2017}
Z.-H. Zhou, ``{A brief introduction to weakly supervised learning},''
  \emph{National Science Review}, vol.~5, no.~1, pp. 44--53, 08 2017.

\bibitem{yang2005review}
J.~Yang, ``Review of multi-instance learning and its applications,''
  \emph{Technical report, School of Computer Science Carnegie Mellon
  University}, 2005.

\bibitem{zhou2006multi}
Z.-H. Zhou, ``Multi-instance learning from supervised view,'' \emph{Journal of
  Computer Science and Technology}, vol.~21, no.~5, pp. 800--809, 2006.

\bibitem{foulds2010review}
J.~R. Foulds and E.~Frank, ``A review of multi-instance learning assumptions,''
  \emph{The Knowledge Engineering Review}, 2010.

\bibitem{Carbonneau_2018}
M.-A. Carbonneau, V.~Cheplygina, E.~Granger, and G.~Gagnon, ``Multiple instance
  learning: A survey of problem characteristics and applications,''
  \emph{Pattern Recognition}, vol.~77, p. 329–353, May 2018.

\bibitem{hickey_NoiseModellingEvaluating_1996}
R.~J. Hickey, ``Noise modelling and evaluating learning from examples,''
  \emph{Artificial Intelligence}, vol.~82, no. 1-2, pp. 157--179, 1996.

\bibitem{frenay_ClassificationPresenceLabel_2014}
B.~Frenay and M.~Verleysen, ``Classification in the {{Presence}} of {{Label
  Noise}}: {{A Survey}},'' \emph{IEEE Transactions on Neural Networks and
  Learning Systems}, vol.~25, no.~5, pp. 845--869, 1994.

\bibitem{lebaher2020}
H.~Le~Baher, V.~Lemaire, and R.~Trinquart, ``On the intrinsic robustness of
  some leading classifiers and symetric loss function - an empiricalevaluation
  (under review),'' \emph{arXiv\string:2010.13570 [cs.LG]}, 2020.

\bibitem{nettleton_StudyEffectDifferent_2010}
D.~F. Nettleton, A.~Orriols-Puig, and A.~Fornells, ``A study of the effect of
  different types of noise on the precision of supervised learning
  techniques,'' \emph{Artificial Intelligence Review}, vol.~33, no.~4, pp.
  275--306, 2010.

\bibitem{folleco_IdentifyingLearnersRobust_2008}
A.~Folleco, T.~M. Khoshgoftaar, J.~Van~Hulse, and L.~Bullard, ``Identifying
  learners robust to low quality data,'' in \emph{2008 {{IEEE International
  Conference}} on {{Information Reuse}} and {{Integration}}}, 2008, pp.
  190--195.

\bibitem{zhu_ClassNoiseVs_2004}
X.~Zhu and X.~Wu, ``Class noise vs. attribute noise: A quantitative study of
  their impacts,'' \emph{Artif. Intell. Rev.}, vol.~22, no.~3, p. 177–210,
  Nov. 2004.

\bibitem{charoenphakdee_symmetric_2019}
N.~Charoenphakdee, J.~Lee, and M.~Sugiyama, ``On symmetric losses for learning
  from corrupted labels,'' in \emph{International Conference on Machine
  Learning}, vol.~97, 2019, pp. 961--970.

\bibitem{Hendrycks2018}
D.~Hendrycks, M.~Mazeika, D.~Wilson, and K.~Gimpel, ``Using trusted data to
  train deep networks on labels corrupted by severe noise,'' in \emph{Advances
  in Neural Information Processing Systems 31}, 2018, pp. 10\,456--10\,465.

\bibitem{Xia2019AreAP}
X.~Xia, T.~Liu, N.~Wang, B.~Han, C.~Gong, G.~Niu, and M.~Sugiyama, ``Are anchor
  points really indispensable in label-noise learning?'' in \emph{NeurIPS},
  2019.

\bibitem{Sukhbaatar2014TrainingCN}
S.~Sukhbaatar, J.~Bruna, M.~Paluri, L.~D. Bourdev, and R.~Fergus, ``Training
  convolutional networks with noisy labels,'' \emph{arXiv: Computer Vision and
  Pattern Recognition}, 2014.

\bibitem{Reed2015TrainingDN}
S.~Reed, H.~Lee, D.~Anguelov, C.~Szegedy, D.~Erhan, and A.~Rabinovich,
  ``Training deep neural networks on noisy labels with bootstrapping,''
  \emph{CoRR}, vol. abs/1412.6596, 2015.

\bibitem{sun_identifying_2007}
J.-w. Sun, F.-y. Zhao, C.-j. Wang, and S.-f. Chen, ``Identifying and
  {Correcting} {Mislabeled} {Training} {Instances},'' in \emph{Future
  {Generation} {Communication} and {Networking} ({FGCN} 2007)}, vol.~1, Dec.
  2007, pp. 244--250, iSSN: 2153-1463.

\bibitem{malossini_detecting_2006}
A.~Malossini, E.~Blanzieri, and R.~T. Ng, ``Detecting potential labeling errors
  in microarrays by data perturbation,'' \emph{Bioinformatics}, vol.~22,
  no.~17, pp. 2114--2121, 2006.

\bibitem{miranda_use_2009}
A.~L.~B. Miranda, L.~P.~F. Garcia, A.~C. P. L.~F. Carvalho, and A.~C. Lorena,
  ``Use of {Classification} {Algorithms} in {Noise} {Detection} and
  {Elimination},'' in \emph{Hybrid {Artificial} {Intelligence} {Systems}}, ser.
  Lecture {Notes} in {Computer} {Science}, 2009, pp. 417--424.

\bibitem{matic_computer_1992}
N.~Matic, I.~Guyon, L.~Bottou, J.~Denker, and V.~Vapnik, ``Computer aided
  cleaning of large databases for character recognition,'' in
  \emph{Proceedings., 11th {IAPR} {International} {Conference} on {Pattern}
  {Recognition}. {Vol}.{II}. {Conference} {B}: {Pattern} {Recognition}
  {Methodology} and {Systems}}, Aug. 1992, pp. 330--333.

\bibitem{van_hulse_knowledge_2009}
J.~Van~Hulse and T.~Khoshgoftaar, ``Knowledge {Discovery} from {Imbalanced} and
  {Noisy} {Data},'' \emph{Data \& Knowledge Engineering}, vol.~68, no.~12, pp.
  1513--1542, Dec. 2009.

\bibitem{Hendrycks2019}
D.~Hendrycks, M.~Mazeika, S.~Kadavath, and D.~Song, ``Using self-supervised
  learning can improve model robustness and uncertainty,'' in \emph{Advances in
  Neural Information Processing Systems 32}, H.~Wallach, H.~Larochelle,
  A.~Beygelzimer, F.~d\textquotesingle Alch\'{e}-Buc, E.~Fox, and R.~Garnett,
  Eds.\hskip 1em plus 0.5em minus 0.4em\relax Curran Associates, Inc., 2019,
  pp. 15\,663--15\,674.

\bibitem{nodet2021importance}
P.~Nodet, V.~Lemaire, A.~Bondu, and A.~Cornu\'ejols, ``Importance reweighting
  for biquality learning,'' in \emph{Proceedings of the International Joint
  Conference on Neural Networks (IJCNN)}, 2021.

\bibitem{PLL1}
E.~H{\"u}llermeier and J.~Beringer, ``Learning from ambiguously labeled
  examples,'' in \emph{Advances in Intelligent Data Analysis VI}, A.~F. Famili,
  J.~N. Kok, J.~M. Pe{\~{n}}a, A.~Siebes, and A.~Feelders, Eds.\hskip 1em plus
  0.5em minus 0.4em\relax Springer Berlin Heidelberg, 2005, pp. 168--179.

\bibitem{zhang2015solving}
M.-L. Zhang and F.~Yu, ``Solving the partial label learning problem: An
  instance-based approach.'' in \emph{IJCAI}, 2015, pp. 4048--4054.

\bibitem{cour2011learning}
T.~Cour, B.~Sapp, and B.~Taskar, ``Learning from partial labels,'' \emph{The
  Journal of Machine Learning Research}, vol.~12, pp. 1501--1536, 2011.

\bibitem{nguyen2008classification}
N.~Nguyen and R.~Caruana, ``Classification with partial labels,'' in
  \emph{Proceedings of the 14th ACM SIGKDD international conference on
  Knowledge discovery and data mining}, 2008, pp. 551--559.

\bibitem{ijcai2019-521}
Q.-W. Wang, Y.-F. Li, and Z.-H. Zhou, ``Partial label learning with unlabeled
  data,'' in \emph{Proceedings of the Twenty-Eighth International Joint
  Conference on Artificial Intelligence, {IJCAI-19}}, 2019, pp. 3755--3761.

\bibitem{JMLR:v12:cour11a}
\BIBentryALTinterwordspacing
T.~Cour, B.~Sapp, and B.~Taskar, ``Learning from partial labels,''
  \emph{Journal of Machine Learning Research}, vol.~12, no.~42, pp. 1501--1536,
  2011. [Online]. Available: \url{http://jmlr.org/papers/v12/cour11a.html}
\BIBentrySTDinterwordspacing

\bibitem{duan2012domain}
L.~Duan, I.~W. Tsang, and D.~Xu, ``Domain transfer multiple kernel learning,''
  \emph{IEEE Transactions on Pattern Analysis and Machine Intelligence},
  vol.~34, no.~3, pp. 465--479, 2012.

\bibitem{ben2010theory}
S.~Ben-David, J.~Blitzer, K.~Crammer, A.~Kulesza, F.~Pereira, and J.~W.
  Vaughan, ``A theory of learning from different domains,'' \emph{Machine
  learning}, vol.~79, no. 1-2, pp. 151--175, 2010.

\bibitem{ennaji2012self}
A.~Ennaji, D.~Mammass, M.~El~Yassa \emph{et~al.}, ``Self-training using a
  k-nearest neighbor as a base classifier reinforced by support vector
  machines,'' \emph{International Journal of Computer Applications}, vol. 975,
  p. 8887, 2012.

\bibitem{torgo20182nd}
L.~Torgo, S.~Matwin, N.~Japkowicz, B.~Krawczyk, N.~Moniz, and P.~Branco, ``2nd
  workshop on learning with imbalanced domains: Preface,'' in \emph{Second
  International Workshop on Learning with Imbalanced Domains: Theory and
  Applications}, 2018, pp. 1--7.

\bibitem{ratner2020snorkel}
A.~Ratner, S.~H. Bach, H.~Ehrenberg, J.~Fries, S.~Wu, and C.~R{\'e}, ``Snorkel:
  Rapid training data creation with weak supervision,'' \emph{The VLDB
  Journal}, vol.~29, no.~2, pp. 709--730, 2020.

\bibitem{varma2018snuba}
P.~Varma and C.~R{\'e}, ``Snuba: Automating weak supervision to label training
  data,'' in \emph{International Conference on Very Large Data Bases}, vol.~12,
  no.~3, 2018.

\bibitem{settles2009active}
B.~Settles, ``Active learning literature survey,'' University of
  Wisconsin-Madison Department of Computer Sciences, Tech. Rep., 2009.

\bibitem{Aggarwal2014}
C.~C. Aggarwal, X.~Kong, Q.~Gu, J.~Han, and P.~S. Yu, ``{Active Learning: A
  Survey},'' in \emph{{Data Classification: Algorithms and Applications}},
  C.~C. Aggarwal, Ed.\hskip 1em plus 0.5em minus 0.4em\relax CRC Press, 2014,
  ch.~22, pp. 571--605.

\bibitem{Santos2014}
D.~Pereira-Santos and A.~C. de~Carvalho, ``{Comparison of Active Learning
  Strategies and Proposal of a Multiclass Hypothesis Space Search},'' in
  \emph{Proceedings of the 9th International Conference on Hybrid Artificial
  Intelligence Systems -- Volume 8480}.\hskip 1em plus 0.5em minus 0.4em\relax
  Springer-Verlag, 2014, pp. 618--629.

\bibitem{Yang2018}
Y.~Yang and M.~Loog, ``{A benchmark and comparison of active learning for
  logistic regression},'' \emph{Pattern Recognition}, vol.~83, pp. 401--415,
  2018.

\bibitem{PereiraSantos2019}
D.~Pereira-Santos, R.~B.~C. Prudêncio, and A.~C. de~Carvalho, ``{Empirical
  investigation of active learning strategies},'' \emph{Neurocomputing}, vol.
  326--327, pp. 15--27, 2019.

\bibitem{hllermeier2019aleatoric}
E.~Hüllermeier and W.~Waegeman, ``{Aleatoric and Epistemic Uncertainty in
  Machine Learning: An Introduction to Concepts and Methods},''
  \emph{arXiv\string:1910.09457 [cs.LG]}, 2019.

\bibitem{Baram2004}
Y.~Baram, R.~El-Yaniv, and K.~Luz, ``{Online Choice of Active Learning
  Algorithms},'' \emph{Journal of Machine Learning Research}, vol.~5, pp.
  255--291, 2004.

\bibitem{Ebert2012}
S.~Ebert, M.~Fritz, and B.~Schiele, ``{Ralf: A reinforced active learning
  formulation for object class recognition},'' in \emph{2012 IEEE Conference on
  Computer Vision and Pattern Recognition}, 2012, pp. 3626--3633.

\bibitem{Hsu2015}
W.-N. Hsu and H.-T. Lin, ``{Active Learning by Learning},'' in
  \emph{Proceedings of the Twenty-Ninth AAAI Conference on Artificial
  Intelligence}.\hskip 1em plus 0.5em minus 0.4em\relax AAAI Press, 2015, pp.
  2659--2665.

\bibitem{Chu2016}
H.-M. Chu and H.-T. Lin, ``{Can Active Learning Experience Be Transferred?}''
  \emph{2016 IEEE 16th International Conference on Data Mining}, pp. 841--846,
  2016.

\bibitem{Collet2018}
T.~Collet, ``{Optimistic Methods in Active Learning for Classification},''
  Ph.D. dissertation, {Universit{\'e} de Lorraine}, 2018.

\bibitem{Pang2018}
K.~Pang, M.~Dong, Y.~Wu, and T.~M. Hospedales, ``{Dynamic Ensemble Active
  Learning: A Non-Stationary Bandit with Expert Advice},'' in \emph{Proceedings
  of the 24th International Conference on Pattern Recognition}, 2018, pp.
  2269--2276.

\bibitem{Konyushkova2017}
K.~Konyushkova, R.~Sznitman, and P.~Fua, ``{Learning Active Learning from
  Data},'' in \emph{Advances in Neural Information Processing Systems 30},
  2017, pp. 4225--4235.

\bibitem{Konyushkova2018}
------, ``{Discovering General-Purpose Active Learning Strategies},''
  \emph{arXiv\string:1810.04114 [cs.LG]}, 2019.

\bibitem{Pang2018b}
K.~Pang, M.~Dong, Y.~Wu, and T.~M. Hospedales, ``{Meta-Learning Transferable
  Active Learning Policies by Deep Reinforcement Learning},''
  \emph{arXiv\string:1806.04798 [cs.LG]}, 2018.

\bibitem{Desreumaux2020LearningAL}
L.~Desreumaux and V.~Lemaire, ``Learning active learning at the crossroads?
  evaluation and discussion,'' in \emph{Workshop Interactive Adaptative
  Learning held at European Conference on Machine Learning}, 2020.

\bibitem{seeger2000learning}
M.~Seeger, ``Learning with labeled and unlabeled data,'' Tech. Rep., 2000.

\bibitem{chapelle_semi-supervised_2006}
O.~Chapelle, B.~Schölkopf, and A.~Zien, Eds.,
  \emph{\BIBforeignlanguage{en}{Semi-supervised learning}}, ser. Adaptive
  computation and machine learning.\hskip 1em plus 0.5em minus 0.4em\relax
  Cambridge, Mass: MIT Press, 2006, oCLC: ocm64898359.

\bibitem{chapelle2009semi}
O.~Chapelle, B.~Schol\-kopf, and A.~Zien, ``Semi-supervised learning,''
  \emph{IEEE Transactions on Neural Networks}, vol.~20, no.~3, pp. 542--542,
  2009.

\bibitem{zhu2005semi}
X.~J. Zhu, ``Semi-supervised learning literature survey,'' University of
  Wisconsin-Madison Department of Computer Sciences, Tech. Rep., 2005.

\bibitem{zhou2010semi}
Z.-H. Zhou and M.~Li, ``Semi-supervised learning by disagreement,''
  \emph{Knowledge and Information Systems}, vol.~24, no.~3, pp. 415--439, 2010.

\bibitem{Bekker_2020}
J.~Bekker and J.~Davis, ``Learning from positive and unlabeled data: a
  survey,'' \emph{Machine Learning}, vol. 109, pp. 719--760, 2020.

\bibitem{khan2014one}
S.~S. Khan and M.~G. Madden, ``One-class classification: taxonomy of study and
  review of techniques,'' \emph{The Knowledge Engineering Review}, vol.~29,
  no.~3, pp. 345--374, 2014.

\bibitem{liu2003building}
B.~Liu, Y.~Dai, X.~Li, W.~S. Lee, and P.~S. Yu, ``Building text classifiers
  using positive and unlabeled examples,'' in \emph{Third IEEE International
  Conference on Data Mining}.\hskip 1em plus 0.5em minus 0.4em\relax IEEE,
  2003, pp. 179--186.

\bibitem{TKDE.2006.16}
G.~P.~C. Fung, J.~X. Yu, H.~Lu, and P.~S. Yu, ``Text classification without
  negative examples revisit,'' \emph{IEEE Trans. on Knowl. and Data Eng.},
  vol.~18, no.~1, p. 6–20, 2006.

\bibitem{blum1998combining}
A.~Blum and T.~Mitchell, ``Combining labeled and unlabeled data with
  co-training,'' in \emph{Proceedings of the eleventh annual conference on
  Computational learning theory}, 1998, pp. 92--100.

\bibitem{davy2005review}
M.~Davy, ``A review of active learning and co-training in text
  classification,'' Trinity College Dublin, Department of Computer Science,
  Tech. Rep., 2005.

\bibitem{ZHAO2017}
J.~Zhao, X.~Xie, X.~Xu, and S.~Sun, ``Multi-view learning overview: Recent
  progress and new challenges,'' \emph{Information Fusion}, vol.~38, pp.
  43--54, 2017.

\bibitem{Mihalcea2004}
R.~Mihalcea, ``Co-training and self-training for word sense disambiguation,''
  in \emph{CoNLL}, 2004.

\bibitem{nigam2000analyzing}
K.~Nigam and R.~Ghani, ``Analyzing the effectiveness and applicability of
  co-training,'' in \emph{Proceedings of the ninth international conference on
  Information and knowledge management}, 2000, pp. 86--93.

\bibitem{Abney02}
S.~P. Abney, ``Bootstrapping,'' in \emph{Proceedings of the 40th Annual Meeting
  of the Association for Computational Linguistics, July 6-12, 2002,
  Philadelphia, PA, {USA}}.\hskip 1em plus 0.5em minus 0.4em\relax {ACL}, 2002,
  pp. 360--367.

\bibitem{Clark03}
S.~Clark, J.~R. Curran, and M.~Osborne, ``Bootstrapping pos-taggers using
  unlabelled data,'' in \emph{Proceedings of the Seventh Conference on Natural
  Language Learning, CoNLL 2003, Held in cooperation with {HLT-NAACL} 2003,
  Edmonton, Canada, May 31 - June 1, 2003}.\hskip 1em plus 0.5em minus
  0.4em\relax {ACL}, 2003, pp. 49--55.

\bibitem{ng-cardie-2003-weakly}
\BIBentryALTinterwordspacing
V.~Ng and C.~Cardie, ``Weakly supervised natural language learning without
  redundant views,'' in \emph{Proceedings of the 2003 Human Language Technology
  Conference of the North {A}merican Chapter of the Association for
  Computational Linguistics}, 2003, pp. 173--180. [Online]. Available:
  \url{https://www.aclweb.org/anthology/N03-1023}
\BIBentrySTDinterwordspacing

\bibitem{Zhou2004Democratic}
Y.~Zhou and S.~A. Goldman, ``Democratic co-learning,'' \emph{16th IEEE
  International Conference on Tools with Artificial Intelligence}, pp.
  594--602, 2004.

\bibitem{Zhou2005TriTraining}
{Zhi-Hua Zhou} and {Ming Li}, ``Tri-training: exploiting unlabeled data using
  three classifiers,'' \emph{IEEE Transactions on Knowledge and Data
  Engineering}, vol.~17, no.~11, pp. 1529--1541, 2005.

\bibitem{saito2017asymmetric}
K.~Saito, Y.~Ushiku, and T.~Harada, ``Asymmetric tri-training for unsupervised
  domain adaptation,'' in \emph{International Conference on Machine Learning},
  2017, pp. 2988--2997.

\bibitem{ruder-plank-2018-strong}
S.~Ruder and B.~Plank, ``Strong baselines for neural semi-supervised learning
  under domain shift,'' in \emph{Proceedings of the 56th Annual Meeting of the
  Association for Computational Linguistics (Volume 1: Long Papers)}, Jul.
  2018, pp. 1044--1054.

\bibitem{li2020dividemix}
J.~Li, R.~Socher, and S.~C.~H. Hoi, ``Dividemix: Learning with noisy labels as
  semi-supervised learning,'' 2020.

\bibitem{zhang2018mixup}
H.~Zhang, M.~Cisse, Y.~N. Dauphin, and D.~Lopez-Paz, ``mixup: Beyond empirical
  risk minimization,'' 2018.

\bibitem{shorten2019survey}
C.~Shorten and T.~M. Khoshgoftaar, ``A survey on image data augmentation for
  deep learning,'' \emph{Journal of Big Data}, vol.~6, no.~1, p.~60, 2019.

\bibitem{qiao2018deep}
S.~Qiao, W.~Shen, Z.~Zhang, B.~Wang, and A.~Yuille, ``Deep co-training for
  semi-supervised image recognition,'' in \emph{Proceedings of the european
  conference on computer vision (eccv)}, 2018, pp. 135--152.

\bibitem{pmlr-v22-urner12}
R.~Urner, S.~B. David, and O.~Shamir, ``Learning from weak teachers,'' in
  \emph{Proceedings of the Fifteenth International Conference on Artificial
  Intelligence and Statistics}, ser. Proceedings of Machine Learning Research,
  vol.~22, pp. 1252--1260.

\bibitem{frenay2013classification}
B.~Fr{\'e}nay and M.~Verleysen, ``Classification in the presence of label
  noise: a survey,'' \emph{IEEE transactions on neural networks and learning
  systems}, vol.~25, no.~5, pp. 845--869, 2013.

\bibitem{weiss2016survey}
K.~Weiss, T.~M. Khoshgoftaar, and D.~Wang, ``A survey of transfer learning,''
  \emph{Journal of Big data}, vol.~3, no.~1, p.~9, 2016.

\bibitem{Conte2010}
D.~{Conte}, P.~{Foggia}, G.~{Percannella}, F.~{Tufano}, and M.~{Vento}, ``A
  method for counting people in crowded scenes,'' in \emph{2010 7th IEEE
  International Conference on Advanced Video and Signal Based Surveillance},
  2010, pp. 225--232.

\bibitem{Charikar2016}
M.~Charikar, J.~Steinhardt, and G.~Valiant, ``Learning from untrusted data,''
  in \emph{Proceedings of the 49th Annual ACM SIGACT Symposium on Theory of
  Computing}, 2017, p. 47–60.

\bibitem{Hataya2019UnifyingSA}
R.~Hataya and H.~Nakayama, ``Unifying semi-supervised and robust learning by
  mixup,'' in \emph{The 2nd Learning from Limited Labeled Data Workshop, ICLR},
  2019.

\bibitem{Gama2014}
\BIBentryALTinterwordspacing
J.~a. Gama, I.~\v{Z}liobaitundefined, A.~Bifet, M.~Pechenizkiy, and
  A.~Bouchachia, ``A survey on concept drift adaptation,'' \emph{ACM Comput.
  Surv.}, vol.~46, no.~4, Mar. 2014. [Online]. Available:
  \url{https://doi.org/10.1145/2523813}
\BIBentrySTDinterwordspacing

\bibitem{dai2007boosting}
W.~Dai, Q.~Yang, G.-R. Xue, and Y.~Yu, ``Boosting for transfer learning,'' in
  \emph{Proceedings of the 24th international conference on Machine learning},
  2007, pp. 193--200.

\bibitem{freund1997decision}
Y.~Freund and R.~E. Schapire, ``A decision-theoretic generalization of on-line
  learning and an application to boosting,'' \emph{Journal of computer and
  system sciences}, vol.~55, no.~1, pp. 119--139, 1997.

\bibitem{hastie2009multi}
T.~Hastie, S.~Rosset, J.~Zhu, and H.~Zou, ``Multi-class adaboost,''
  \emph{Statistics and its Interface}, vol.~2, no.~3, pp. 349--360, 2009.

\bibitem{caruana1997multitask}
R.~Caruana, ``Multitask learning,'' \emph{Machine learning}, vol.~28, no.~1,
  pp. 41--75, 1997.

\bibitem{patrini2017making}
G.~Patrini, A.~Rozza, A.~Menon, R.~Nock, and L.~Qu, ``Making deep neural
  networks robust to label noise: a loss correction approach,'' in \emph{IEEE
  Conference on Computer Vision and Pattern Recognition (CVPR)}, 2017.

\bibitem{4803844}
L.~{Bruzzone} and M.~{Marconcini}, ``Domain adaptation problems: A dasvm
  classification technique and a circular validation strategy,'' \emph{IEEE
  Transactions on Pattern Analysis and Machine Intelligence}, vol.~32, no.~5,
  pp. 770--787, 2010.

\bibitem{fang2020rethinking}
T.~Fang, N.~Lu, G.~Niu, and M.~Sugiyama, ``Rethinking importance weighting for
  deep learning under distribution shift,'' 2020.

\bibitem{sugiyama2010density}
M.~Sugiyama, T.~Suzuki, and T.~Kanamori, ``Density ratio estimation: A
  comprehensive review (statistical experiment and its related topics),'' 2010.

\bibitem{huang2007correcting}
J.~Huang, A.~Gretton, K.~Borgwardt, B.~Sch{\"o}lkopf, and A.~J. Smola,
  ``Correcting sample selection bias by unlabeled data,'' in \emph{Advances in
  neural information processing systems}, 2007, pp. 601--608.

\bibitem{gretton2009covariate}
A.~Gretton, A.~Smola, J.~Huang, M.~Schmittfull, K.~Borgwardt, and
  B.~Sch{\"o}lkopf, ``Covariate shift by kernel mean matching,'' \emph{Dataset
  shift in machine learning}, vol.~3, no.~4, p.~5, 2009.

\end{thebibliography}
\end{document}